# The Combined Technique for Detection of Artifacts in Clinical Electroencephalograms of Sleeping Newborns

Vitaly Schetinin and Joachim Schult

*Abstract*—In this paper we describe a new method combining the polynomial neural network and decision tree techniques in order to derive comprehensible classification rules from clinical electroencephalograms (EEGs) recorded from sleeping newborns. These EEGs are heavily corrupted by cardiac, eye movement, muscle and noise artifacts and as a consequence some EEG features are irrelevant to classification problems. Combining the polynomial network and decision tree techniques, we discover comprehensible classification rules whilst also attempting to keep their classification error down. This technique is shown to outperform a number of commonly used machine learning technique applied to automatically recognize artifacts in the sleep EEGs.

## I. INTRODUCTION

Electroencephalograms (EEGs) representing the weak potentials invoked by the brain activity give EEG-experts objective information for analysis and classification (e.g., [1] – [6]). Although EEGs are noise and nonstationary signals varying from one patient to the other in a large range of amplitudes and frequencies, the EEGs have been used to assist clinicians to diagnose such diseases as apnea, Alzheimer, dementia and schizophrenia (e.g., [5] – [7]). To make reliable decisions clinicians have to properly separate neural activity of patients from EEG artifacts caused by electrode noise, eye movement, cardiac, and muscle activities. To do so, they use such methods as independent component analysis (e.g., [1], [8]), regression methods (e.g., [9], [10]), and principle component analysis (e.g., [11]).

In sleep research when EEGs are analysed in frequency domain muscle artifacts can be automatically recognised by a threshold technique comparing the high frequency activity of brief segments and the local background activity calculated in window of a predefined length [e.g., 12]. Similarly to other outlying methods this technique discards brief 4-second EEG segments in which the power of a high frequency band is out of a threshold dependent on the average value calculated in 3-min window.

The methods [3] – [6] suggested for classification of clinical EEGs are based on the fully connected feed-forward neural network (FNN) for which users have to properly predefine a suitable network structure as well as a learning method for fitting synaptic weights of the network. Although such FNNs can learn to classify EEGs well, corruptions of EEGs still lead to ambiguous results because the underlining brain and muscle activities share their characteristics such as wave shape and frequencies [5], [6]. In the meantime, and just as important, the classification models learnt by FNNs cannot be comprehensible for medical experts due to a large number of synaptic connections [13] – [19].

In contrast to the FNNs with a predefined structure, a Group Method of Data Handling (GMDH) allows us to induce well-suited neural networks from data [20] – [22]. GMDH algorithms generate a multilayer neural network step-by-step by growing up new layers of neurons. The network grows until a predefined criterion reaches a minimum located near to a global one. This criterion is based on a cross-validation error function assuming training and validation data subsets. The resultant classification model is described by a concise set of the neurons with a given transfer function, for example, a polynomial one.

To derive classification rules from data, we could directly apply the Decision Tree (DT) technique which exploits a greedy or hill-climbing strategy [13]. This technique is capable of partitioning data well, however in the presence of irrelevant and noise features it may derive classification rules with poor generalization ability [15], [16]. This drawback of the DT technique is particularly overcome by using some pruning strategies described in [13], [14].

In this paper we describe a new combined technique developed for learning artifact recognition in clinical EEGs recorded from sleeping newborns. In this research we use fruitful machine learning and pattern recognition methods to induce new rules for an automated recognition of EEG

Manuscript received June 3, 2003; revised August 6, 2003 and November 14, 2003. This work was supported by the University of Jena, Germany, and particularly by the University of Exeter, UK, under EPSRC Grant GR/R24357/01.

V. Schetinin is with the Department of Computer Science, the University of Exeter, Exeter, EX4 4QF, UK (phone: +44 139-226-2090; fax: +44 139-226-4067; e-mail: V.Schetinin@ex.ac.uk).

J. Schult was with TheorieLabor, the University of Jena, Ernst-Abbe-Platz 4, Jena, D-07740, Germany (e-mail: Joachim_Schult@web.de).

artifacts and then compare their performances. Within the techniques that are commonly used for analyzing sleep EEGs in frequency domain (e.g., [5], [12]), for inducing these rules we do not use channels besides EEG. However, the experts labeling the EEG segments used additional information coming from other channels which were useful for visual recognition of the EEG artifacts. So the cardiac, eye movement, muscle, and noise artifacts visually recognizable in the EEGs were labeled.

Within our research we also evaluate how well the discovered rules can recognize the artifacts in the EEGs recorded from newborns of which background neural activity varies during sleep hours. Such variations heavily affect the accuracy of artifact recognition [12]. In this research, however, we do not use the average estimations of background neural activity as suggested in [12] and attempt learning a recognition rule exploiting the spectral features calculated for the current EEG segment.

In the framework of our technique first we learn the GMDH-type neural network from the given training data. This network has a nearly minimal number of neurons and involves those features which make the most important contribution to classification of the patterns. Then using the selected features and the training examples that have been correctly classified, we induce an appropriate DT. As a result we derive a comprehensible classification rule whilst also attempting to keep its classification error down.

Below in section 2 we describe the classification problem and clinical EEG data used in our experiments. In section 3 we briefly describe GMDH algorithms and present our learning algorithm. Then in sections 4 and 5 we present the DT induction technique and compare the performance of our approach on the clinical EEGs. Finally we refer to related work and briefly discuss the obtained results.

## II. THE CLASSIFICATION PROBLEM AND METHOD

Referring to [7], when recording from newborns, EEGs should be analyzed with a particular attention to the presence and type of eye movements, facial movements, respiration (regular or irregular), sucking, crying etc. Extracerebral monitors are needed in routine recordings, including at least electrooculogram (EOG), respiration rate measurement, and electrocardiogram (ECG). Only a reduced number of scalp electrodes, generally never more than the set in a 16-channel recording, are applicable.

Active sleep, the antecedent of rapid eye movement sleep, is usually indicated by irregular respiratory patterns with interspersed, brief episodes that often precede clusters of eye movements. Contrary to adult physiology, prominent, subtle motor activity, especially of the face (e.g., grimacing, smiling), accompanies this state.

The sleep spindling samples are slower in frequency and more anteriorly distributed in newborns compared with older infants. These infrequently appear at the beginning of quiet sleep as rudimentary, immature, asymmetric, and asynchronous 10-16 Hz EEG waveforms.

In our experiments the clinical EEGs were recorded from sleeping newborns via the standard EEG channels, C3 and C4, sampled with 100 Hz. These EEGs were transformed to the frequency domain by using a fast Fourier transform as described in [5]. The spectral features were calculated at the 10-second segments into six frequency bands such as subdelta (0-1.5 Hz), delta (1.5-3.5 Hz), theta (3.5-7.5 Hz), alpha (7.5-13.5 Hz), beta 1 (13.5-19.5 Hz), and beta 2 (19.5-25 Hz). Additionally for each band the values of relative and absolute powers were calculated. Such values were calculated for channels C3 and C4 as well as for their sum, C3+C4, so the total number of the features was 36. Values of these features were normalized to be with zero mean and unit variance.

Using the additional channels, the EEG-experts have recognized cardiac, eye movement, muscle and noise artifacts and labeled all the EEG segments recorded from 42 newborns aged between 36 and 51 weeks. In our first experiment two EEGs recorded from two newborns were available, one for learning and the other for testing the classification rule. The training and testing EEGs contain 1347 and 808 labeled segments in which the artifact rates are 6.53% and 8.79%, respectively. Note that all these segments were labeled by one EEG-expert.

In our second experiment 40 EEG records were available: the 20 records containing 17,094 segments were randomly selected for training and the remaining 20 records containing 21,250 segments were used for testing. The artifact rates in the training and testing datasets were 20.7% and 35.6% respectively. These segments were labeled by two experienced EEG-experts which applied their subjective strategies of the artifact recognition. In our experiments we did not identify the EEG records labeled by these experts. So taking in account a high artifact rate and the differences in strategies of the experts, we can expect that a vector of class labels is much more noise than in the first experiment.

In our experiments we focus mainly on comparing the performance of our technique and the commonly used machine learning methods applied to the artifact recognition in clinical EEG presented in the frequency domain. In this paper we do not study on other methods that analyze the additional channels (e.g., EOG and ECG) in the time domain.

The idea behind our method of EEG classification is to combine the GMDH-type neural network and decision tree techniques. The first technique is used to induce a well-suited neural network which involves the relevant features and allows us to find the misclassified training examples. Then using the decision tree technique we derive an accurate classification rule from the training data cleaned from the irrelevant features and misclassified examples. After such cleaning the training data are presented by the relevant variables and the boundaries between the classes become smooth. So under these circumstances we can expect an increase in the chance to find a decision tree that will better generalize than an original GMDH-type network.

### III. GMDH-Type Polynomial Networks

In this section we briefly describe GMDH algorithms which are capable of inducing well-suited polynomial neural networks from data. The induced GMDH-type networks can generalize well because their size or complexity is nearly minimal.

#### A. GMDH Algorithms

GMDH-type polynomial networks are the multilayered feed-forward networks consisting of the so-called supporting neurons which have at least two inputs $v_1$ and $v_2$ [20] – [22]. A transfer function $g$ of the neurons is described by short-term polynomials, for example, by a nonlinear polynomial:

$$y = g(\mathbf{v}; \mathbf{w}) = w_0 + w_1 v_1 + w_2 v_2 + w_3 v_1 v_2, \qquad (1)$$

where $\mathbf{v} = (v_1, v_2)$ is an input vector and $\mathbf{w} = (w_0, w_1, \ldots, w_3)$ is a weight vector of the neuron consisting of coefficients $w_0$, $w_1$, …, $w_3$.

Using the supporting neurons, GMDH algorithm builds new generation, or layer, of the candidate-neurons and then selects the best of them. The candidate-neurons are selected with the so-called exterior criterion which evaluates the generalization ability of neurons on the unseen data defined as the validation data. So user has to predefine the exterior criterion as well as the number of neurons, $F$, selected in each layer. Giving the large $F$, the user increases the chance to find out a global minimum of cross-validation error however the large values of $F$ increases the computational expenses. In practice GMDH algorithms perform enough well for $F$ equal to the number of input variables, $m$. The best performance is achieved for $F = 0.4 * \binom{m}{2}$, where $\binom{m}{2}$ is the number of combinations by 2 from $m$ given inputs [20].

The layers of GMDH-type networks grow up one-by-one. In the first layer the candidate-neurons are connected to inputs $x_1$, …, $x_m$, and in the next layer they are connected to the outputs of the $F$ neurons selected in the previous layer. Some GMDH algorithms allow also combining between the input nodes and neurons taken from the previous layers [23].

As a given activation function (1) has two arguments $v_1$ and $v_2$, the first layer, $r = 1$, consists of candidate-neurons $y_1^{(1)}$, …, $y_{L_1}^{(1)}$, where $L_1 = \binom{m}{2}$. Having training these neurons, GMDH algorithm selects the $F$ best of them and then generates the next layer in which it generates $L_r = \binom{F}{2}$ candidate-neurons. This procedure of generation and selection of the candidate-neurons cycles and the network grows in size while the value of the exterior criterion decreases.

The exterior criterion can be defined on the training and validation datasets $\mathbf{D}_A$ and $\mathbf{D}_B$ consisting of $n_A$ and $n_B$ examples respectively, $n_A + n_B = n$, $\mathbf{D}_A \cup \mathbf{D}_B = \mathbf{D}$, where $\mathbf{D} = (\mathbf{X}, \mathbf{y}^o)$, $\mathbf{X}$ is a $n \times m$ matrix of the input data and $\mathbf{y}^o$ is a $n \times 1$ target vector. The training data $\mathbf{D}_A$ are used to fit the weight vector $\mathbf{w}$ of the supporting neuron so that to minimize the sum squared error, $e = \sum_k (g(\mathbf{v}^{(k)}; \mathbf{w}) - y_k^o)^2, k = 1,\ldots,n_A$. The validation data $\mathbf{D}_B$ are used to control the complexity or the number of layers of the GMDH-type network during learning.

For fitting weights $\mathbf{w}$ the conventional GMDH algorithms exploit a least square method which provides the effective estimates of weights if the training data are Gaussian distributed [20] – [22]. For real-world data for which a Gaussian distribution is unrealistic such estimations become biased [24]. One way used in such cases to avoid this problem is to make the estimations of weights without unrealistic assumptions about the distribution function of training data; one such learning method is described below in section 3.2.

The complexity of the GMDH-type network is controlled by calculating the value of the exterior criterion, $CR_i^{(r)}$, $i = 1, \ldots, L_r$, for each candidate-neuron on the whole data $\mathbf{D}$ as follows:

$$CR_i^{(r)} = \sum_k (g_i(\mathbf{v}^{(k)}; \mathbf{w}) - y_k^o)^2, k = 1,\ldots,n. \qquad (2)$$

The value of $CR_i^{(r)}$ as we can see is dependent on how well the $i$th neuron classifies the unseen data $\mathbf{D}_B$. Therefore the value of $CR^{(r)}$ is expected large for the neurons with poor generalization ability and small for the neurons which generalize well.

The values of $CR^{(r)}$ are calculated for all the candidate-neurons in the layer $r$ and GMDH then sorts them in an ascending order, $CR_{i1}^{(r)} \leq \ldots \leq CR_F^{(r)} \leq \ldots \leq CR_{Lr}^{(r)}$, so that the first $F$ neurons provide the best classification accuracy. The minimal value of the exterior criterion, $CR_m^{(r)}$, equal to $CR_{i1}^{(r)}$ is used to check the following stopping rule:

$$\left| CR_m^{(r)} - CR_m^{(r-1)} \right| < \Delta, \qquad (3)$$

where $\Delta > 0$ is a constant given by user.

This rule is based on an observation that value of $CR_m^{(r)}$ decreases rapidly at the first layers of GMDH-type network and relatively slowly near to an optimal number of layers, and further increasing the number of layers causes increasing the value of $CR_m^{(r)}$ because of over-fitting [20] – [22]. Thus the number of layers in the network increases one-by-one until the stopping rule is met at the layer $r^*$. Subsequently we can take a desired GMDH-type network of a nearly optimal complexity from the ($r^* - 1$)th layer.

#### B. Fitting the Neuron Weights

As mentioned above unrealistic assumptions on the distribution of real-world data lead to the biased estimations the neuron weights. However we can use a learning algorithm which is not dependent on the distribution of training data. Below we describe our method.

Accordingly to the given transfer function (1), the inputs of the supporting neurons are connected to the pairs of the input variables $(x_i, x_j)$, $\forall i \neq j = 1, \ldots, m$ for the first layer and to the outputs of the neurons $(y_i, y_j)$, $\forall i \neq j = 1, \ldots, F$, for the next layers. So for the training and validation of the supporting neurons, we can denote their input data as the $n_A \times 2$ and $n_B \times 2$ matrices $\mathbf{U}_A$ and $\mathbf{U}_B$, respectively. Using these notations we can describe our learning method as follows.

Initially $k$ is set to zero and the algorithm initiates a weight vector $\mathbf{w}^0$ by random values. At the next step $k$ the algorithm calculates a $n_A \times 1$ error vector, $\mathbf{\eta}_A^k$, on the data $\mathbf{D}_A$ as follows:

$$\mathbf{\eta}_A^k = g(\mathbf{U}_A; \mathbf{w}^k) - \mathbf{y}_A^o. \qquad (4)$$

On the validation data $\mathbf{D}_B$, it calculates the $n_B \times 1$ error vector $\mathbf{\eta}_B^k$:

$$\mathbf{\eta}_B^k = g(\mathbf{U}_B; \mathbf{w}^k) - \mathbf{y}_B^o, \qquad (5)$$

as well as the corresponding mean squared error, $e_B(k)$, of the neuron:

$$e_B(k) = 1/n_B \sum_i (\eta_{Bi}^k)^2, i = 1,...,n_B. \qquad (6)$$

The error $e_B$ has to be minimized during learning for a finite number of steps $k$. Formally we can complete the learning if the following inequality is met at the step $k^*$:

$$e_B(k^* - 1) - e_B(k^*) < \delta, \qquad (7)$$

where $\delta > 0$ is a constant which depends on the level of noise in data $\mathbf{X}$ as well as on the ratio $n_B/n$ given by user.

Until this inequality is met, the current weight vector $\mathbf{w}^{k-1}$ is updated accordingly with the following learning rule:

$$\mathbf{w}^k = \mathbf{w}^{k-1} - \chi \|\mathbf{U}_A\|^{-2} \mathbf{U}_A \mathbf{\eta}_A^{k-1}, \qquad (8)$$

where $\chi$ is a given learning rate, and $\|\cdot\|$ is a Euclidean norm.

The desired estimation of weights is achieved for a finite number of steps, $k^*$, if the learning rate $\chi$ lies between 1 and 2, for a proof see [24]. In our experiments the best performance was obtained with a ratio $n_B/n = 0.5$, a learning rate $\chi = 1.9$, $\delta = 1.5 \cdot 10^{-2}$ and the initial weights distributed by the Gaussian $N(0, 1)$; in this case the number $k^*$ did not exceed 30 steps.

## IV. THE DECISION TREE ALGORITHM

The idea behind the DT algorithm used in our experiments for rule extraction is similar to C4.5 algorithm described in [13]. First we define the training subsets $\mathbf{X}_0$ and $\mathbf{X}_1$ consisting of $n_0$ and $n_1$ examples assigned to classes 0 and 1, $n_0 + n_1 = n$, where $n$ is now the number of the training examples correctly classified by GMDH-type network. This network exploits $m$ input variables $x_1, …, x_m$ presenting the $\mathbf{X}_0$ and $\mathbf{X}_1$, all these variables are relevant to the classification problem.

Let us also initialize a decision tree, $\mathbf{T}$, and define a procedure find_node that is invoked with the $\mathbf{X}_0$ and $\mathbf{X}_1$ as the parameters. This procedure searches for an input variable, $v_1$, and a threshold, $q_1$, which provide the best partition of the subsets $\mathbf{X}_0$ and $\mathbf{X}_1$. A new node $f(v_1, q_1)$ involving variable $v_1$ and threshold $q_1$ is added to the $\mathbf{T}$. The procedure find_node calls itself while splitting nodes contain more than $p$ given training examples belonging to classes 0 and 1. The main steps of this procedure are:

1. Search a threshold $q_i$ of node $f(x_i, q_i)$ for each variable $x_1, …, x_m$.
2. Find a feature $v_1$ which divides the subsets $\mathbf{X}_0$ and $\mathbf{X}_1$ with a minimal error.
3. Create a new node $f(v_1, q_1)$ and add it to the $\mathbf{T}$.
4. Calculate the outputs of the node $f(v_1, q_1)$.
5. Find the correctly classified, $A_0$ and $A_1$, and misclassified, $A_{10}$ and $A_{01}$, examples of the $\mathbf{X}_0$ and $\mathbf{X}_1$.
6. If $A_{10}$ contains more than $p$ examples, then find_node($\mathbf{X}_0(A_0, :), \mathbf{X}_1(A_{10}, :)$).
7. If $A_{01}$ contains more than $p$ examples, then find_node($\mathbf{X}_0(A_{01}, :), \mathbf{X}_1(A_1, :)$).

As a result, the variable $\mathbf{T}$ contains a desired DT. This DT is capable of generalizing well, because the training data $\mathbf{X}_0$ and $\mathbf{X}_1$ were beforehand cleaned from the misclassified examples and irrelevant features.

For large-scale data the search for threshold $q$ through all the training examples may take a long time. To reduce the computational expenses we applied the technique [20] which draws the random values of $q$ for a given number of attempts. So giving a rational number of attempts, we can achieve a good performance for an acceptable time. Below we apply our technique to the clinical EEGs.

## V. EXPERIMENTAL RESULTS

In this section we describe two experiments aimed at evaluating the performance of our technique and some machine learning methods that are commonly used on EEGs. These experiments were carried out with the labeled EEGs recorded from several newborns as described in section 2, the first with 2 EEGs and the second with 40 EEGs.

In the experiments we compared such machine learning techniques as $k$-nearest neighbor ($k$-nn), C4.5 DT, the standard FNN, and the GMDH described in section 3 with our polynomial neural network (PNN) and combined (PNN&DT) techniques. As the artifact rate in the EEGs was between 6.53% and 35.6%, additionally to the performance we evaluated the sensitivity and specificity of the classifiers. The sensitivity is calculated as TP/(TP + FN) and the specificity as TN/(TN + FP), where TP, TN, FP, and FN are the number of patterns classified as true positive, true negative, false positive and false negative, respectively. The performance is calculated as (TP + TN)/(TP + TN + FP + FN). Here positive patterns are associated with artifacts and negative with normal segments.

Searching for the parameters of DT splitting nodes, we used the technique described in [13] which tests a given number $\lambda$ of values drawn from a uniform distribution ranged between the minimal and maximal values of the feature tested by the splitting node. For the first experiment we have set the $\lambda$ equal to 300 and for the second equal to 150.

For training the FNNs, we used the standard technique based on the principal component analysis and a fast Levenberg-Marquardt back-propagation learning algorithm provided by MATLAB. The FNNs with a given number of hidden neurons and randomly initialized weights were trained 30 times.

Inducing the GMDH-type networks, we used a modified algorithm described in [23] which allows the random connections between different layers of neurons. This algorithm takes a random pair of the neurons and creates a new neuron which is added to the network if $\mu_{new} < \min(\mu_1, \mu_2)$, where $\mu_{new}$, $\mu_1$, and $\mu_2$ are the values of criterion (2) calculated for the new and taken neurons, respectively. The growth of the network terminates after a prespecified number of failed attempts of improving the performance; this number was specified equal to 7. For GMDH-type networks as well as for the PNNs, we used a transfer function (1) and $F$ equal 240.

In our first experiment we used one EEG containing 1347 labeled segments for training and other EEG containing 808 segments for testing. The artifacts rates in these EEGs were 6.53% and 8.79%, respectively. Table I lists the mean and 95% confidence interval of the sensitivity, specificity, and performance calculated for the DT, FNN, GMDH, PNN, as well as for the PNN&DT on the testing EEG.

TABLE I
PERFORMANCE OF CLASSIFIERS ON THE 808 TESTING EEG SEGMENTS

| # | Classifier | Sensitivity, % | Specificity, % | Performance, % |
|---|---|---|---|---|
| 1 | DT | 68.9±5.0 | 98.4±2.2 | 95.8±2.4 |
| 2 | FNN | 57.5±37.4 | 99.3±1.4 | 95.7±3.2 |
| 3 | GMDH | 63.1±8.0 | 99.9±0.2 | 96.6±0.8 |
| 4 | PNN | 63.1±13.4 | 99.5±1.0 | 96.3±1.8 |
| 5 | PNN&DT | 65.8±16.4 | 98.9±2.6 | 96.0±2.4 |

For inducing the DTs, we used the pruning strategy cutting the nodes splitting fewer than 5 data points or 0.4% of the training data. The resultant DTs have correctly classified 95.8±2.4% of the testing EEG segments. The number of nodes in these DTs was 8.4±1.8.

Applying the standard neural network technique, we found out that the FNNs exploiting 10 hidden neurons and 24 principal components provided the best performance Over 30 runs the FNNs correctly classified 95.7±3.2% of the testing EEG data.

The GMDH-type networks with the settings described above have correctly classified 96.6±0.8% of the testing EEG data. The PNNs have nearly the same performance and correctly classified 96.3±1.8% of the testing data.

The best PNN for which the performance was 97.4% consists of seven neurons connected with the following eight features:

- *AbsPowSubdeltaC3*, the absolute power of subdelta in channel C3,
- *AbsPowSubdeltaC4*, the absolute power of subdelta in channel C4,
- *RelPowThetaC4*, the relative power of theta in C4,
- *RelPowTheta*, the relative power of theta in C3+C4,
- *AbsPowThetaC4*, the absolute power of theta in C4,
- *RelPowAlphaC4*, the relative power of alpha in C4,
- *AbsPowAlpha*, the absolute power of alpha in C3+C4, and
- *RelPowBeta2C3*, the relative power of beta2 in C3.

Defining a transfer function (1) as a polynomial function $y = P(v_1, v_2; [w_0\ w_1\ w_2\ w_3])$ in which arguments $v_1$ and $v_2$ are connected either to outputs of the previous neurons or to the features listed above, the best PNN can be comprehensively described by a following set of the seven polynomials:

$y_1 = P(AbsPowThetaC4, RelPowThetaC4;\ [0.9466\ -0.0875\ 0.0731\ 0.0703])$,
$y_2 = P(AbsPowSubdeltaC3, RelPowBeta2C3;\ [0.9335\ -0.1309\ -0.0656\ -0.0648])$,
$y_3 = P(AbsPowSubdeltaC4, RelPowTheta;\ [0.9325,\ -0.2036,\ -0.0076,\ 0.0028])$,
$y_4 = P(AbsPowAlpha, RelPowAlphaC4;\ [0.9295\ -0.1931\ 0.0337\ 0.0362])$,
$y_5 = P(y_1, y_2;\ [0.1886\ -0.5950\ 0.6661\ 0.7637])$,
$y_6 = P(y_3, y_4;\ [0.2500\ -0.0032\ -0.5401\ 1.3314])$,
$y_7 = P(y_5, y_6;\ [0.2823\ -0.1038\ 0.0455\ 0.7832])$,

Following our technique, we used the induced PNNs to remove the irrelevant features and the misclassified examples from the training data. The cleaned data were then used to induce the DT as described in section 4. Over 30 runs the performance of the DTs was 96.0±2.4% and the number of DT splitting nodes ranging between 1 and 4 was 2.4±2.4.

The best DT for which the performance was 97.3% exploits two features which are *AbsPowSubdelta*, the absolute power of subdelta in C3+C4, and *AbsPowBeta1C3*, the absolute power of beta1 in C3. A diagram of this DT is depicted below.

*AbsPowSubdelta* < 0.7027: artifact(0.0)
*AbsPowSubdelta* ≥ 0.7027:
 *AbsPowSubdelta* < 1.2813:
  *AbsPowBeta1C3* < 0.6718: artifact(0.0)
  *AbsPowBeta1C3* ≥ 0.6718: artifact(0.8571)
 *AbsPowSubdelta* ≥ 1.2813: artifact(0.9726)

Here artifact($p_i$) is referred to the frequency probability $p_i$ of an artifact output for the $i$th splitting node, that is $p_i = \dfrac{n_1^{(i)}}{n_1^{(i)} + n_2^{(i)}}$, where $n_1^{(i)}, n_2^{(i)}$ are the numbers of training examples at the $i$th splitting node labeled as EEG artifact and normal segments, respectively.

This experiment shows that removing the misclassified training examples and the irrelevant features we could reduce the size of the DT and keep its classification error down. The average size of the DTs induced by our technique decreased from 8.4 to 2.4 while the average performance slightly increased from 95.8% to 96.0%. Such DTs are easily interpreted by experts.

Note that in our first experiment the artifacts rates in the two EEGs used for training and testing were relative small. Both EEGs were labeled by one expert applying one strategy of artifact recognition. So it would be interesting to check the performance of the induced DTs on other EEGs recorded from new patients and labeled by two EEG-viewers.

The performance of the induced DTs was evaluated on the 21,250 EEG segments recorded from 20 newborns. Its values were 68.6±3.0% and the sensitivity and specificity were 42.5±13.6% and 83.0±7.0%, respectively. So we can see that the DTs derived from one EEG are not able to classify EEGs recorded from new patients well. This means that the classification model is dependent on patient specifics of which influence might be diminished by inducing DTs from a large set of EEG records.

In our second experiment we used much more EEG data: for training 17,097 and for testing 21,250 EEG examples recorded from 40 newborns as described in section 2. The artifact rates in these EEGs were also much higher: 20.7% in the training and 35.6% in the testing EEG data.

In this experiment the best performance of *k*-nn was achieved for 5 neighbors and 24 principal components. The values of its sensitivity, specificity, and performance were 62.6%, 74.2%, and 71.1%, respectively.

The performances of the DT, FNN, GMDH as well as the PNN and PNN&DT averaged over 30 runs on the testing data are presented in Table II.

TABLE II
PERFORMANCE OF CLASSIFIERS ON THE 21,250 TESTING EEG SEGMENTS

| # | Classifier | Sensitivity, % | Specificity, % | Performance, % |
|---|---|---|---|---|
| 1 | DT | 61.4±11.8 | 74.5±7.6 | 69.8±2.6 |
| 2 | FNN | 57.3±11.2 | 78.1±8.6 | 70.7±3.8 |
| 3 | GMDH | 52.1±9.6 | 84.9±4.8 | 73.2±1.2 |
| 4 | PNN | 53.4±14.6 | 84.1±8.8 | 73.2±2.6 |
| 5 | PNN&DT | 51.4±14.2 | 85.5±8.4 | 73.5±2.8 |

For inducing the DTs we used the searching strategy drawing $\lambda = 150$ random values from a uniform distribution and cut of the splitting nodes containing fewer 100 data points or 0.6% of the training data. The resultant DTs have correctly classified 69.8±2.6% of the testing EEG segments and the number of splitting nodes was 48.5±9.4.

The best performance of the FNN was achieved for 30 hidden neurons and 24 principal components. These FNNs have correctly classified 70.7±3.8% of the testing data, this is only 0.9% higher than that for the DTs.

Running the GMDH-type neural networks, we achieved the performance 73.2±1.2%. The values of their sensitivity and specificity were 52.1±9.6% and 84.9±4.8%, respectively.

The PNNs correctly classified 73.2±2.6% of the testing EEG segments. The values of their sensitivity and specificity were 53.4±14.6% and 84.1±8.8%, respectively.

Having applied our PNN&DT technique, we could slightly improve the performance to 73.5±2.8% and achieve the sensitivity and specificity equal to 51.4±14.2% and 85.5±8.4%, respectively. The size of the DTs averaged over 30 runs was 5.0±3.8.

We can see that the DT induced by our PNN&DT technique is almost 10 times shorter than that induced by the conventional C4.5 technique. In the meantime our DTs outperform C4.5 DTs with a 95% confidence interval between 4.3% and 2.9% and the significance $\alpha < 10^{-14}$. So we conclude that our PNN&DT technique performs on the EEG data better than C4.5 DT technique.

Analyzing the induced DTs, we found out two DTs providing the best performance equal to 76.5%. These DTs exploit one feature *AbsPowBeta2*, the absolute power of beta2 in C3+C4.

The sensitivity and specificity of the first DT were 49.4% and 91.4%, respectively. A diagram of this DT is depicted below.

*AbsPowBeta2* < -0.0205: artifact(0.0041)
*AbsPowBeta2* ≥ -0.0205:
   *AbsPowBeta2* < 0.1765: artifact(0.1687)
   *AbsPowBeta2* ≥ 0.1765: artifact(0.9313)

The sensitivity and specificity of the second DT were 50.9% and 90.6%, respectively. This DT consisting of one node is depicted as the following diagram.

*AbsPowBeta2* < 0.1599: artifact(0.0045)
*AbsPowBeta2* ≥ 0.1599: artifact(0.9965)

Observing the results in Table II we can conclude that our technique certainly outperforms the C4.5 and neural-network techniques on the sleep EEGs. However analyzing the induced DTs described above, we can see that both exploit only one feature *AbsPowBeta2* presenting the power of a high frequency band mentioned in [12]. That is, these DTs testing one feature *AbsPowBeta2* band without the information about background neural activity of sleeping newborns cannot perform enough well.

## VI. RELATED WORK

EEG correction techniques focus mainly on removing ocular artifacts from the EEG and removing artifacts caused by muscle activity, cardiac signals, and electrode noise. Regression methods for removing muscle noise are impractical because signals from multiple muscle groups require different reference channels [8].

Several methods proposed for removing eye-movement artifacts are based on regression in the time domain [25] – [27]. However, in the time domain removing eye artifacts tends to overcompensate for blink artifacts [28], [29].

Regression in the frequency domain [9], [10] can account for frequency-dependent transfer function differences from EOG to EEG. Regression methods in the time or frequency domains depend on a regressing channel.

In [30] a method of eye artifact removal has been proposed using a spatiotemporal dipole model that requires a priori assumptions about the number of dipoles for saccade, blink, and other eye movements. A technique [11] has been proposed for removing ocular artifacts using principal component analysis. Corrected EEG data could be obtained by removing corrupted components by the simple inverse computation. Using this technique, ocular artifacts can be removed more effectively than by regression or by using spatiotemporal dipole models.

Makeig *et al*. [1] proposed an approach to the analysis of EEG data based on independent component analysis (ICA). The ICA algorithm can be used to separate neural activity from muscle and blink artifacts in spontaneous EEG data. ICA methods are based on the assumptions that the signals recorded on the scalp are mixtures of time courses of temporally independent cerebral and artifact sources, that potentials arising from different parts of the brain, scalp, and body are summed linearly at the electrodes, and that propagation delays are negligible. Corrected EEG signals can be derived by eliminating the contributions of the artifact sources.

Several methods suggested for an automated analysis of EEGs are based on neural networks [2] – [6]. To deal with artifacts the system [6] identifies the type of EEG corruptions and characterizes the ocular and muscle artifacts. The valuable spectral features of EEGs were extracted by using parametric modeling and cross-correlation.

Breidbach *et al*. [5] have classified the EEGs recorded from sleeping newborns by using a FNN including 72 inputs and two output neurons all with a sigmoid activation function. The learning algorithm which they used aims to maximize a Euclidean distance between the output vectors belonging to different classes. Analyzing the clusters in a space of two output variables, they discovered the correlation between the neuron outputs and the risk groups and conclude that a neural network with five inputs produce a better classification accuracy.

The rule extraction techniques described in [15] – [18] prune the synaptic weights and retrain FNNs. Such a strategy is based on a trade-off between the complexity and the classification accuracy of the extracted rules, and additionally can be computationally expensive. The other rule extraction method [19] is based on successive regularization and retraining the FFNs. The resultant rules extracted by this method are dependent on the given parameters of regularization and training. This method is also computationally expensive.

The GMDH has been suggested to learn polynomial models of a nearly minimal complexity from data [20] – [22]. Such models can be comprehensively described by a set of short-term polynomials which users may find more observable than the fully connected neural-network classifiers.

VII. CONCLUSION

Although fully connected as well as polynomial neural networks can learn to classify EEGs well, such classifiers cannot be comprehensible for clinicians. On the other hand the rule extraction and decision tree techniques which are able to produce the comprehensible rules are commonly based on the trade-off between the complexity and the classification accuracy of rules. Contrary to this approach we combine the polynomial neural network and the decision tree techniques to be able keeping the classification error down.

Using our technique in this paper we have learnt a new rule for an automated recognition of cardiac, eye movement, muscle and other artifacts in sleep EEGs, presented in frequency domain. For recognition we assumed a rule exploiting the features calculated for the current EEG segment while additional information coming from other channels outside EEG has been used only by the experts to label the EEG artifacts.

As a result the discovered rule is easily interpreted as decision tree testing the power of a high frequency band. An analogous feature is tested for recognizing muscle artifacts in sleep EEG by using the threshold technique [12]. Additionally this technique evaluates the background neural activity at 3-minute window adjusted for adult EEGs. However in our experiments this information has not been used for learning the new rule. So the first practical result of our research is that the discovered recognition rule is comprehensible for EEG-experts and that this rule matches well with the rule described in [12]. The second practical issue is that we compared the performance of commonly used machine learning methods on sleep EEGs without using the additional information coming from channels besides EEG. As a final point, the new rule has been learnt for an automated recognition of all the types of EEG artifacts including cardiac, eye movement, muscle, and electrode noise which were visually recognizable for the EEG-experts.

Having comparing the classification accuracy on the testing EEG data, it can be seen that our technique outperforms the commonly used machine learning methods. However, because of a large variation in the EEGs of sleeping newborns, we could not achieve a high accuracy of artifact recognition without additional information about the background neural activity as described in [12].

Thus we can conclude that combining the polynomial neural network and decision tree techniques allow us provide the comprehensible rules and at the same time keep the classification error down. We believe that this rule extraction technique seems promising for applications to clinical EEGs.

ACKNOWLEDGMENT

The authors are grateful to Frank Pasemann for enlightening discussions, Joachim Frenzel and Burkhart Scheidt from the Clinic of the University of Jena for the clinical EEG records, and to Jonathan Fieldsend from the University of Exeter for useful comments.

**Vitaly Schetinin** received a Ph.D. degree from the Penza State University, Russia, in 1996.

He has worked as an Associate Professor of Computer Science at the Penza State University and later as a Research Fellow at the University of Jena, Germany, and the University of Exeter, UK. His research interests include machine learning, pattern recognition, and neural network methods and their applications.

**Joachim Schult** graduated from the University of Kiel and received a Ph.D. from the University of Hamburg in 1980.

He was a Lecturer of Biology at the University of Hamburg and later worked as a Research Associate at the EEG laboratory headed by Prof. E. Basar at the University of Lübeck. He was also a Head of the TheorieLabor at the University of Jena. His research interests involve phylogeny and evolution of Arthropoda, the structure and function of electroencephalogram, biosemiotics and theoretical concepts in biology.